\documentclass{article}
\usepackage{ijcai25}

\usepackage{times}
\usepackage{soul}
\usepackage{url}
\usepackage[hidelinks]{hyperref}
\usepackage[utf8]{inputenc}
\usepackage[small]{caption}
\usepackage{graphicx}
\usepackage{amsmath}
\usepackage{amsthm}
\usepackage{booktabs}
\usepackage[switch]{lineno}

\usepackage{amsmath}
\usepackage{amsfonts}       % blackboard math symbols
\usepackage{mathtools}
\usepackage{dsfont}
\usepackage[noend]{algpseudocode}
\usepackage{algorithm}

\usepackage{threeparttable}
\usepackage{multirow}
\usepackage{multicol}
\usepackage{booktabs}  
\usepackage{xcolor} 
\usepackage{colortbl}
\usepackage{subfigure}
\usepackage{amssymb}
\title{Enhancing Sampling Protocol for Point Cloud Classification Against Corruptions}

 \author{
 Chongshou Li$^1$  \and
 Ping Tang$^1$\and
 Xinke Li\footnote{Corresphonding author}$^{2}$\and
 Yuheng Liu$^3$ \And 
 Tianrui Li$^1$\\
 \affiliations
$^1$School of Computing and Artificial Intelligence, Southwest Jiaotong University\\
$^2$College of Computing, City University of Hong Kong\\
$^3$SWJTU-Leeds Joint School, Southwest Jiaotong University\\
\emails
\{lics, trli\}@swjtu.edu.cn,
tangpin1874@163.com,
xinkeli@cityu.edu.hk,
yuhengliu02@gmail.com
} %\fi

\begin{document}
\maketitle

\begin{abstract}
Established sampling protocols for 3D point cloud learning, such as Farthest Point Sampling (FPS) and Fixed Sample Size (FSS), have long been relied upon. However, real-world data often suffer from corruptions, such as sensor noise, which violates the benign data assumption in current protocols. As a result, these protocols are highly vulnerable to noise, posing significant safety risks in critical applications like autonomous driving. To address these issues, we propose an enhanced \underline{point} cloud \underline{s}ampling \underline{p}rotocol, \textbf{\underline{PointSP}}, designed to improve robustness against point cloud corruptions. PointSP incorporates key point reweighting to mitigate outlier sensitivity and ensure the selection of representative points. It also introduces a local-global balanced downsampling strategy, which allows for scalable and adaptive sampling while maintaining geometric consistency. Additionally, a lightweight tangent plane interpolation method is used to preserve local geometry while enhancing the density of the point cloud. Unlike learning-based approaches that require additional model training, PointSP is architecture-agnostic, requiring no extra learning or modification to the network. This enables seamless integration into existing pipelines. Extensive experiments on synthetic and real-world corrupted datasets show that PointSP significantly improves the robustness and accuracy of point cloud classification, outperforming state-of-the-art methods across multiple benchmarks.
%To address these challenges, we propose an enhanced point cloud sampling protocol, PointSP, which consists of two key components: 1) key point sampling for robust feature selection, and 2) full point resampling for flexible sample size adjustment. Differentiated strategies are applied during the training and inference phases. Specifically, a local-global balanced downsampling method is designed to perform random key point selection during training and mitigate noise during inference. In addition, a local-geometry-preserving upsampling technique is incorporated, enabling flexible sample size during training and compensating for missing data during inference. Importantly, the proposed protocol does not require changes to model architecture or additional training, making it easy to integrate with existing systems. Despite its simplicity, PointSP significantly enhances the robustness of point cloud learning, as demonstrated by its superior performance over state-of-the-art methods across multiple benchmarks for corrupted point cloud classification.
\end{abstract}

\section{Introduction}
In the rapidly evolving field of 3D data perception via deep learning \cite{qi2017pointnet,qi2017pointnet++,guo2020deep}, point cloud sampling serves as a critical component in the standard learning and recognition pipeline~\cite{hu2020randla,qian2022pointnext,yu2022point,zhang2022point}. 
Following the legacy of pioneer works~\cite{qi2017pointnet,qi2017pointnet++}, existing sampling protocols are primarily designed and optimized for clean data, without taking into account corruptions. However, due to the high complexity of real-world, point cloud data are almost always incomplete and with noise in practice~\cite{ren2022pointcloud}, posing threats to  3D deep learning applications. For example, noisy background points or slight perturbations generated by inaccurate processing or sensor error can significantly decrease the deep model performance \cite{ren2022pointcloud,sun2022benchmarking}.  Such performance drops can lead to serious safety consequences, especially in critical 3D applications like autonomous driving. Therefore, it is necessary to rethink and redesign point cloud sampling protocols with a focus on robustness against corruptions to ensure reliable 3D deep learning in real-world conditions. 

\begin{figure}[t]
\centering
\includegraphics[width=0.49\textwidth]{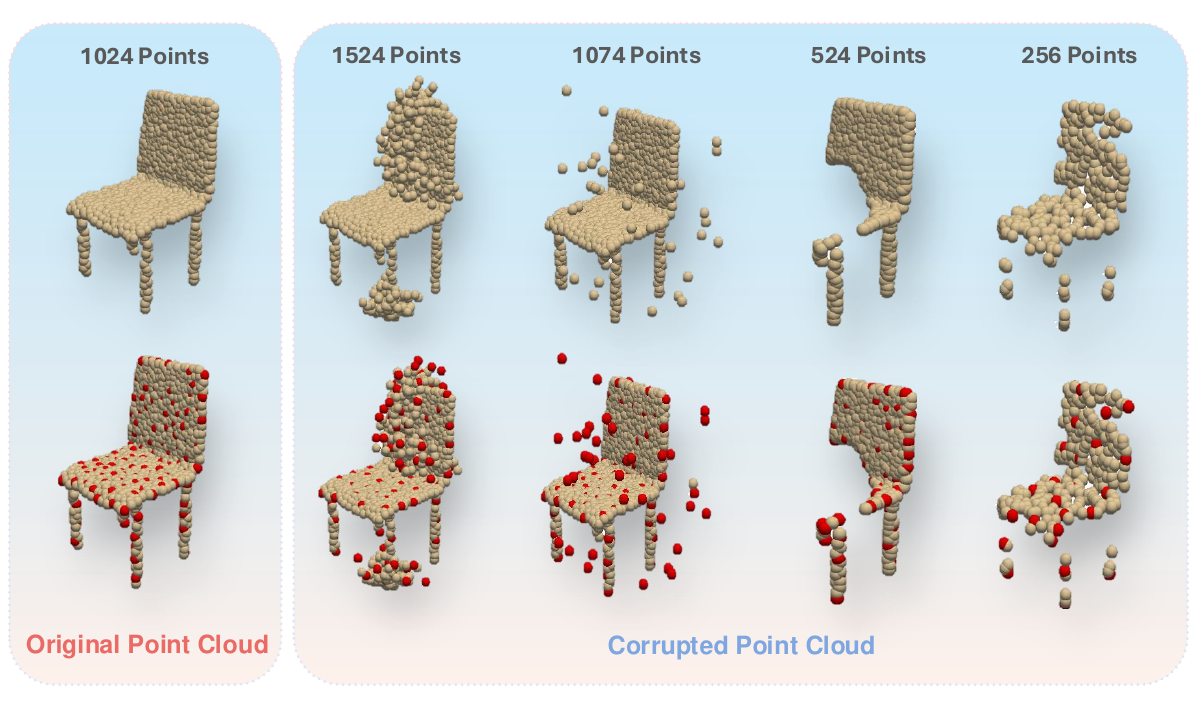}
\caption{The first row presents the original point clouds, while the second row highlights the sampled key points, with those selected by farthest point sampling (FPS) shown in red. The existing standard sampling protocol is not optimized for corrupted point cloud in practice. As a result, a standard-trained PointNet classifies them as \textcolor{green}{Chair}, \textcolor{red}{Vase}, \textcolor{red}{Table}, \textcolor{red}{Mantel}, and \textcolor{red}{Sofa}.
Towards this issue, we propose to enhance the protocol by revising FPS into \textbf{new key points selections} and integrating \textbf{full points resampling} into process.}
\label{fig:teaser}
\end{figure}

One crucial limitation of established sampling protocols is that they are sub-optimal under the corrupted data distribution. For instance, the protocol samples a fixed number of points in data preparation, namely, Fixed Sample Size (FSS)~\cite{qi2017pointnet}. This convention overlooks the facts that point clouds in the real world naturally vary in size and density. These varied sizes are even obvious in particular corruptions such as occlusions and density-related noise~\cite{sun2022benchmarking,ren2022pointcloud}.   
The misclassfication results on various corrupted data are illustrated in Figure \ref{fig:teaser}.
Another aspect is that the widely used Farthest Point Sampling (FPS)~\cite{eldar1997farthest} for key points selection is especially vulnerable to outliers 
due to its inherent basis of Euclidean distance and sensitivity to sparse points~\cite{yan2020pointasnl}. 
Several works have considered updating a specific step to deal with this issue, like %AdaptPoint \cite{wang2023sample},  
PointASNL \cite{yan2020pointasnl} and ADS \cite{hong2023attention}.  The learning-based methods put extra effort into module training and may be potentially overfitting. Overall, none of them propose a comprehensive and alternative solution to overcome the sampling protocol limitations. 

To overcome these limitations, we propose an enhanced point cloud sampling protocol, \textbf{PointSP},  
by revising key points selection %the \textbf{\underline{D}ownsampling} process 
and  full points resampling.% and %\textbf{\underline{R}esampling} process 
% before inputting, 
 The implementation of the proposed protocol involves randomizing the sampling during training and processing noisy point clouds during inference. To achieve this, the key point sampling process assisted by point reweighting is applied to ensure that potential outliers are not captured. The point weight is named as isolation rate evaluating the extent of local isolation for a point. Moreover, the proposed full points resampling randomizes the sample size during training; and restores insufficient point clouds in the inference stage. We achieve an local-global-balanced downsampling offering a continuous spectrum between local and global sampling. Inspired by shape-invariant perturbation \cite{huang2022shape}, we realize an lightweight upsampling via a tangent plane interpolation technique that enhances the density of point cloud data while preserving local geometry.Overall, the enhanced sampling protocol is learning-free thus straightforward to implement and can be seamlessly integrated into the existing point cloud analysis pipeline.

Our contributions are summarized as follows: 
\begin{itemize}
    \vspace{-1mm}
    \item We first comprehensively revisit the long-existing sample protocol for point cloud learning through the lens of data corruption. Based on the analysis, we propose an alternative protocol to enhance the robustness of point cloud learning.
    \item  We develop three learning-free techniques as the key of protocol, %namely,  
    point reweighting, tangent plane interpolation and local-global balanced sampling, which can deal with point cloud corruption in different aspects. The whole proposed protocol is free of model architecture change and extra learning, thus it can be implemented to replace the current protocol with minimal pains and fits almost all 3D deep models. 
    % \vspace{-1mm}
    \item Extensive experiments are conducted on synthesis and real corrupted 3D point cloud datasets. The results have demonstrated that the proposed protocol is able to improve the robustness of 3D point cloud classification and outperform the latest methods. 
\end{itemize}

\section{Related Work} \label{sec:lr}
%\textcolor{red}{Citation formats need to be fixed.}
\textbf{Point Cloud Sampling.} 
Point cloud sampling techniques typically consist of: 1) downsampling, also known as ``simplification'' \cite{dovrat2019learning}, and 2) upsampling \cite{zhang2022point}. These techiniques are divided into non-learning-based and learning-based methods \cite{zhang2022point}. Traditional non-learning-based downsampling techniques include Farthest Point Sampling (FPS) \cite{eldar1997farthest}, Random Sampling (RS) \cite{hu2020randla}, Poisson Disk Sampling (PDS) \cite{ying2013intrinsic}, and voxelization \cite{lv2021approximate}. Conversely, learning-based downsampling methods account for downstream tasks \cite{dovrat2019learning,lang2020samplenet,qian2020pugeo,qian2023task}. Upsampling is categorized into learning-based \cite{yu2018pu,yu2018ec,dai2020neural,qiu2022pu,he2023grad,he400grad} and non-learning-based approaches \cite{alexa2003computing,huang2013edge,wu2015deep}. The non-learning-based sampling techniques are particularly susceptible to outliers due to their inherent structural limitations; meanwhile, the learning-based methods are either also sensitive to noise or dependent on downstream tasks and prone to overfitting.  This paper introduces simple yet effective point cloud sampling techiniques to overcome these challenges.

\textbf{ Point Cloud  Classification.} PointNet \cite{qi2017pointnet} has been a trailblazer in utilizing deep learning for point cloud analysis, with notable extensions such as PointNet++ \cite{qi2017pointnet++}, GDANet \cite{xu2021learning}, Point Transformer (PCT) \cite{guo2021pct},  and CurveNet \cite{xiang2021walk}. However, the performance of these models significantly deteriorates with corrupted real-world data \cite{uy2019revisiting,ren2022pointcloud,sun2022benchmarking}. To tackle this issue, existing literature offers three main types of solutions. The first focuses on modifying the model by altering its structure or training strategies, such as pooling operations based on sorting \cite{sun2020adversarial} and model aggregation \cite{dong2020self}. The second type includes certified methods, exemplified by Pointguard, which theoretically enhances model robustness through certified classification \cite{liu2021pointguard}. The third type is data-driven approaches that directly cleanse corrupted data, with notable methods including IF-defense \cite{wu2020if} and DUP-Net \cite{zhou2019dup}. This paper aims to advance robustness from a new perspective by refining point cloud sampling protocol during data preparation, while the non-trivial architecture modification is avoided. 

\textbf{Point Cloud Data Augmentation.} Point cloud augmentation is a widely recognized practice in the deep learning community, employed to improve the generalization capabilities of neural networks. Traditional augmentation methods, including random scaling, rotation, and jitter, are somewhat limited in their effectiveness for point cloud analysis \cite{zhu2024advancements}. Recent advancements have introduced sophisticated techniques such as PointCutMix~\cite{zhang2022pointcutmix}, PointAugment \cite{li2020pointaugment}, PointMixup \cite{chen2020pointmixup}, and PointWOLF \cite{kim2021point}. However, they suffer from various limitations. For instance, while PointMixup \cite{chen2020pointmixup} and PointWOLF \cite{kim2021point} largely rely on predefined transformations, PointAugment \cite{li2020pointaugment} emphasizes global transformations, often at the expense of local geometric details. To our knowledge, the sampling augmentation of point cloud for robust classification has been largely unexplored. In this work, we aim to enhance sampling protocols specifically tailored for robust point cloud classification, addressing this critical gap.

\section{Proposed Sampling Protocol} 
\subsection{Existing Sampling Protocol}
\begin{figure*}[t]
\centering
\includegraphics[width=1.0\textwidth]{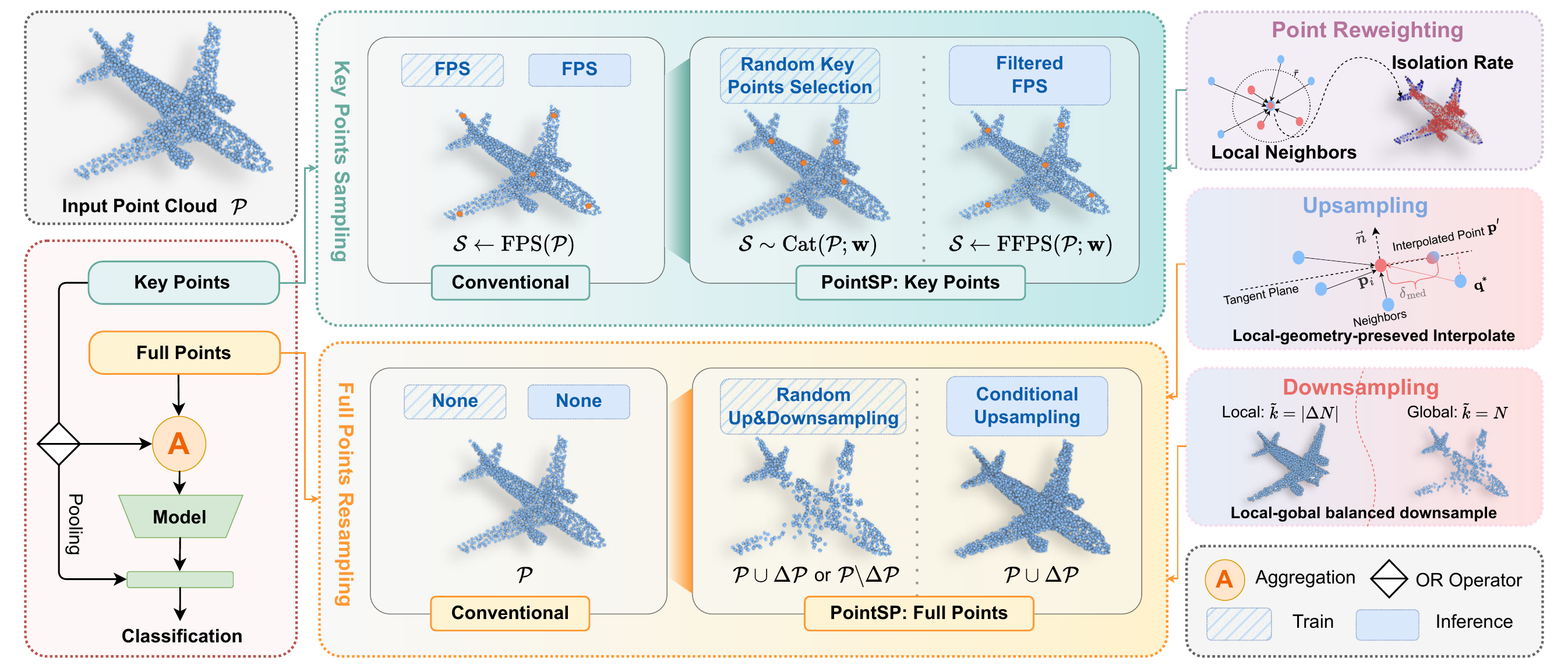}
\caption{PointSP: enhanced protocol of point cloud sampling for robust classification. The existing and conventional protocol used farthest point sampling (FPS) and non-processed points for input. 
In our protocol, randomized key point sampling and full points resampling (random up\& downsampling) are used in training to conduct sampling-based data augmentation. During inference, filtered FPS (FFPS) is implemented to bypass outliers, and an upsampling strategy is used to densify sparse input. We propose the concept of isolation rate, the upsampling by tangent plane interpolation and the local-global balanced downsampling to obtain point weights and resampled points, respectively. }
\label{fig:main}
\vspace{1em}
\end{figure*}

Mainstream 3D point cloud classification frameworks employ a standardized protocol predicated on processing clean point clouds of predetermined size, e.g., 1,024. Formally, the input point cloud is defined as $\mathcal{P} = \{\mathbf{p}_i\}^N_{i=1}$, where each point $\mathbf{p}_i \in \mathbb{R}^{3}$ and $N$ represents a fixed size. A common technique of this protocol is the Farthest Point Sampling (FPS) algorithm, which identifies key points that serve as anchors for subsequent feature aggregation or pooling operations as in \cite{qi2017pointnet++}. FPS iteratively constructs a subset $\mathcal{S}\subseteq \mathcal{P}$ by selecting points $\mathbf{s}_t$ according to:
\begin{equation}
\mathbf{s}_t = \arg \max_{\mathbf{p}_i\in \mathcal{P}} \min_{\mathbf{s}\in\mathcal{S}} \|\mathbf{p}_i - \mathbf{s} \|_2,
\end{equation}
where $t=1,\cdots, |\mathcal{S}|$ denotes the iteration index.

This conventional protocol, however, exhibits significant limitations when confronted with real-world scenarios where point clouds are invariably corrupted and of variable size. We can characterize a corrupted point cloud $\mathcal{P'}$ through the following formulation:
\begin{equation}\label{eq:noisy_points}
    \mathcal{P'} = \mathcal{P}\backslash \mathcal{P}_s\cup \mathcal{O},
\end{equation}
where $\mathcal{P}_{s}$ represents points removed from the original clean cloud $\mathcal{P}$, while $\mathcal{O}$ comprises introduced noise points or outliers, typically arising from occlusion effects and sensor imperfections. This formulation exposes three critical vulnerabilities in the current protocol:

\begin{itemize}

    \item FPS inherently selects outliers ($\mathcal{O}$) as key points due to its distance-based criterion.

    \item Missing points ($\mathcal{P}_s$) lead to information loss without compensatory mechanisms.

   \item The variable size of $\mathcal{P}_s$ and $\mathcal{O}$ violate the fixed-size input constraint.
\end{itemize}

Recent empirical studies \cite{ren2022pointcloud,sun2022benchmarking} have confirmed these shortcomings, showing substantial performance degradation under corruption scenarios. This necessitates a fundamental redesign of the sampling protocol to enhance robustness in point cloud processing.

\subsection{New Sampling Protocol}
We propose the enhanced and learning-free sampling protocol towards \textbf{key points selection} and \textbf{full points preprocessing}, respectively. The new point cloud sampling protocol (PointSP) is visualized in Figure~\ref{fig:main}.

\textbf{Key Points Sampling.} We propose distinct sampling strategies for training and inference stages to select key points. \underline{During training}, we employ a stochastic sampling approach based on point-wise weights by $
\mathbf{s}_t \sim \text{Cat}(\mathcal{P};\mathbf{w})$, 
where $\text{Cat}(\cdot;\cdot)$ denotes the categorical distribution, and $\mathbf{w}=\{w_1,\cdots,w_N\}$ represents the point weights. These weights are derived from the isolation rate of each point (detailed in Section \ref{sec:3.3}), with more isolated points receiving lower weights to reduce their sampling probability.
\underline{During inference}, we introduce Filtered FPS (FFPS), a deterministic sampling method that modifies the conventional FPS by incorporating binary weights:
\begin{equation}
\mathbf{s}_t = \arg \max_{\mathbf{p}_i\in \mathcal{P}} \hat{w}_i \min_{\mathbf{s}\in\mathcal{S}_t} \|\mathbf{p}_i - \mathbf{s} \|_2.
\end{equation}
where $\hat{w}_i$ is a binary weight determined by thresholding $w_i$ at the $\omega$-th quantile (typically $\omega=0.95$) of all weights. This effectively filters out the most isolated points, which are often outliers or noise. The binarization simplifies implementation while maintaining effectiveness.

\textbf{Full Points Resampling.} Our resampling protocol differentiates training and inference stages. \underline{During training}, we deliberately introduce size variations to enhance model robustness against real-world point clouds often with non-uniform sizes. Given an input point cloud $\mathcal{P}$ with $N$ points, we randomly adjust its size to $N + \Delta N$, where the random variable $\Delta N$ can be either positive (upsampling) or negative (downsampling). The modified point cloud $\tilde{\mathcal{P}}$ is obtained through:
\begin{equation}\label{eq:up-downsample}
\tilde{\mathcal{P}} = \begin{cases}
\mathcal{P} \cup \Delta \mathcal{P}, & \text{if } \Delta N > 0 \\
\mathcal{P} \setminus \Delta \mathcal{P}, & \text{if } \Delta N \leqslant 0
\end{cases}
\end{equation}
\underline{During inference}, our protocol ensures consistent point cloud size by applying conditional upsampling. Specifically, for any input point cloud $\mathcal{P}$ with insufficient points (i.e., $|\mathcal{P}| < N$), we supplement it with additional points via upsampling to reach the target size $N$, namely, $\tilde{\mathcal{P}} = \mathcal{P} \cup \Delta \mathcal{P}$. The specific techniques for generating additional points ($\Delta \mathcal{P}$) in upsampling and downsampling are detailed in Section~\ref{sec:3.3}.

\subsection{Proposed Sampling Techniques}\label{sec:3.3}
We introduce three key techniques of the new sampling protocol, namely,  designed downsampling and upsampling techinques for full points resampling, and the point reweighting techinque for key point sampling.

\textbf{Point Reweighting.} The point-wise weight in key points sampling can be defined by the concept of \textit{Isolation Rate}. At first, we calculate the radius of a sphere containing $k$ nearest neighbors of each point in $\mathcal{P}$, which is given by $
    r_i= \max_{\mathbf{q}_j\in \mathcal{N}_i^k}\left \| \mathbf{p}_i-\mathbf{q}_j \right \|_2$
, where $\mathcal{N}_i^k \subseteq \mathcal{P}$ is the set of $k$ neighbors of $i$-th point $\mathbf{p}_i$. We further define \textit{Isolation Rate} for each point as $w_i$, given by,
\begin{align} \label{eq:weight}
      w_i = \textup{Pr}_{d\in \mathcal{D}_i}(d \geqslant \bar{r}), \mathcal{D}_i = \{\|\mathbf{q}_j-\mathbf{p}_i\|_2: \forall \mathbf{q}_j\in\mathcal{N}^k_i\}
\end{align}
where $\bar{r}=\text{Median}(\{r_i\}_{i=1}^N)$ is the median of all radius and $\textup{Pr}_{Y}(X)$ is the probability of $X$ given condition $Y$. The isolation rate of a point suggests the extent of a point being isolated, \textit{i.e.,} far from others in a probability way. Although a few associated concepts were proposed to calculate the exact local radius of points \cite{sotoodeh2006outlier} and identify outliers,  the isolation rate is naturally fit for point weighting due to the probability representation.

\textbf{Downsampling: Local-global-balanced downsampling.} For $\Delta N \leqslant 0$ in ~\eqref{eq:up-downsample}, we remove points through a flexible neighborhood-based approach. Specifically, we randomly select a center point $\mathbf{p}_i$ and consider its $\tilde{k}$ nearest neighbors $\mathcal{N}^{\tilde{k}}_i$, where $\tilde{k}$ is randomly sampled from $[|\Delta N|, N]$. The points to be removed are selected according to:
\begin{equation}
\Delta \mathcal{P} \sim \{\Gamma: \Gamma \subseteq \mathcal{N}^{\tilde{k}}_i, |\Gamma| = -\Delta N\}
\end{equation}
This strategy offers a continuous spectrum between local and global downsampling: when $\tilde{k}=|\Delta N|$, it removes a concentrated local patch; when $\tilde{k}=N$, it performs global random sampling; and intermediate values of $\tilde{k}$ provide balanced local-global downsampling. This flexibility helps simulate various real-world point cloud corruptions.

 \textbf{Upsampling: Local-geometry-preserved Interpolation.}  For $\Delta N > 0$, we propose a interpolation method that generates new points. For each point, we perform interpolation between it and a randomly selected neighbor on their shared tangent plane, as detailed in Algorithm \ref{alg:estimate_perb}. This approach ensures the preservation of local geometric features, particularly surface normals, while increasing point density. The upsampled points are sampled from the interpolation set $\hat{\mathcal{P}}$ by
\begin{equation}
\Delta \mathcal{P} \sim \{\Gamma: \Gamma \subseteq \hat{\mathcal{P}}, |\Gamma| = \Delta N\}.
\end{equation}
This interpolation-based strategy effectively balances point cloud densification with geometric preservation, resulting in more natural and structurally coherent augmented data.

\textbf{Computational Cost.} We design the sampling protocol in the way that poses minimal computational effort beyond the original protocol. Particularly, the implementation of FPS in conventional protocol involves the calculation of point paired distances with complexity of $O(N^2)$. The proposed point reweighting and interpolation (for all points) techinques can utilize the same paired distances and induce extra operations with $O(kN)$ complexity. Such extra computation effort is minor since $k\ll N$. More experimental results of computational costs are presented in Appendix.
\begin{algorithm}  
\caption{Local Geometry Preserved Interpolation}
\label{alg:estimate_perb}  
\begin{algorithmic}[1]  
\Require Point cloud $\mathcal{P}$, Point cloud normal $\{\mathbf{n}_i\}_{i=1}^N$ , A query point $\mathbf{p}_i \in \mathcal{P}$, Integer $k$
\Ensure A new point $\hat{\mathbf{p}}_i$
\State $\mathcal{N}^k \gets \text{$k$NN}(\mathcal{P}, \mathbf{p}_i, 
k)$ \Comment{$k$ nearest neighbors of $\mathbf{p}_i$}  
\State $\delta_{\text{med}} \gets \text{Median}(\{ \|\mathbf{p}_i - \mathbf{q}\| : \mathbf{q} \in \mathcal{N}^k \})$ \Comment{Compute the median of local distances as interpolation norm.}  
\State $\mathbf{q}^* \sim \mathcal{N}^k$ \Comment{Sample a random neighbor}  
%\State $\vec{v} \gets q - p$ \Comment{Compute the vector from $p$ to $q$}  
\State $\mathbf{v}_i \gets (\mathbf{I}-\mathbf{n}_i\mathbf{n}_i^{\top})\cdot(\mathbf{q}^* - \mathbf{p}_i)$ \Comment{Compute the interpolation direction on tangent plane.}  
\State $\hat{\mathbf{p}}_i \gets \mathbf{p}_i + \delta_{\text{med}} \frac{\mathbf{v}_i}{\| \mathbf{v}_i \|}$ \Comment{Create interpolated point }
\State \textbf{return} $\hat{\mathbf{p}}_i.$  
\end{algorithmic} 
\end{algorithm}  
% \vspace{-1.5em}

\section{Experimental Studies} \label{sec:es}

\subsection{Experimental Setup} 

\textbf{Dataset and Model.} We utilize models trained on ModelNet40~\cite{wu20153d} to conduct experiments on three corrupted datasets: ModelNet40-C, PointCloud-C, and OmniObject-C. The ModelNet40-C~\cite{sun2022benchmarking} and PointCloud-C~\cite{ren2022pointcloud} are datasets applying 15 and 7 distinct corruptions to ModelNet40's test set, totaling 2,468 objects. The OmniObject-C, based on OmniObject3D~\cite{wu2023omniobject3d}, has real-scanned 362 objects corrupted by the methods proposed in~\cite{ren2022pointcloud}. For 3D deep models, we employ PointNet~\cite{qi2017pointnet}, PointNet++~\cite{qi2017pointnet++}, GDANet~\cite{xu2021learning}, CurveNet~\cite{xiang2021walk}, PCT~\cite{guo2021pct}, following the pipeline in ModelNet40-C including batch size and training protocol. We note that all experiments are run on NVIDIA GeForce RTX 3090 GPUs.

\textbf{Parameters Setting.} The number of nearest neighbors $k$ used in point weight computation is set to 20, which follows the common setup. During the inference phase, the key points sampling applies a threshold $\omega$ of 0.95, exploring the learning as depicted in Figure \ref{fig:w-fps-omega}, meaning that FFPS filters out points within the lowest 5\% of point weights.

\textbf{Evaluation Protocol.} 
We report the error rates (ER) and mean error rates (mER) across multiple corruptions on the three corrupted datasets for performance evaluation. 
A smaller ER indicates a superior performance. More implementation details are referred to Appendix.
\vspace{-0.5em}
\subsection{Main Results}

\textbf{Overall Results.} Mean error rates (mERs) for the three corrupted datasets are presented in Table \ref{tab:OA_ER}. To facilitate a comprehensive comparison, we include multiple baseline models. The results clearly indicate that the proposed PointSP significantly enhances PCT and CurveNet; mERs decrease by approximately 10\% across all datasets, with the most substantial improvement observed in PointCloud-C.

\begin{table}[htbp]
  \centering
  \small 
  \begin{tabular}{l|ccc}
    \toprule
     % Model(\%)$\downarrow$ 
      Method & MNC  &  PCC  & OmniC\\
    \midrule \midrule
    PointNet \cite{qi2017pointnet}          &28.3 &33.7 &65.2\\
    PointNet++ \cite{qi2017pointnet++}        &30.6 &27.7 &73.9\\
    DGCNN \cite{wang2019dynamic}             &25.9 & 23.5 & 73.7 \\
    RSCNN  \cite{liu2019relation}            &26.2 &26.1 & 72.4\\
    CurveNet \cite{xiang2021walk}         &23.1 &24.4 &67.9 \\
    SimpleView \cite{goyal2021revisiting}        &27.2 & 24.3& 71.8\\
    GDANet \cite{xu2021learning}            &23.5 &24.6 & 70.9 \\
    PCT   \cite{guo2021pct}     &25.5 &25.8 & 69.8\\
    \midrule
    PCT+\textbf{PointSP}            &\textbf{15.8} &\textbf{12.5} &60.8\\
    CurveNet+\textbf{PointSP}       &\textbf{15.8} &13.7 &\textbf{57.9}\\
    \bottomrule
%    \toprule[1]
  \end{tabular}
  \caption{Mean error rate (mER) across all corruptions of popular 3D deep models w/o our protocol on three datasets. The best mERs are highlighted in bold. MNC, PCC and OmniC represent ModelNet40-C, PointCloud-C and OmniObject-C, respectively.}
  \label{tab:OA_ER}
\vspace{-1em}
\end{table}

\textbf{Results on ModelNet40-C.}
Extensive evaluations of PointSP on ModelNet40-C utilizing five 3D deep models revealed its superiority. Compared to five enhancement techniques (CutMix-R~\cite{zhang2022pointcutmix}, CutMix-K~\cite{zhang2022pointcutmix}, Mixup~\cite{chen2020pointmixup}, Rsmix~\cite{lee2021regularization}, and PGD~\cite{sun2021adversarially}), PointSP significantly improved all models. Across multiple corruption types, PointSP consistently achieved the lowest error rates: 24.1\% for ``\textit{Density}'', 9.5\% for ``\textit{Noise}'', and 11.1\% for ``\textit{Transform}'', demonstrating robustness. Notably, its unique randomized size sampling in resampling and FFPS in downsampling effectively tackled ``\textit{Density}'' and ``\textit{Noise}'' corruptions, enhancing resilience and eliminating outliers, respectively. Detailed corruption results are provided in the Appendix.

%\textbf{Results on PointCloud-C.}   Table \ref{tab:merged_ER} presents the results on the PointCloud-C dataset, emphasizing the robustness of PointSP in enhancing point cloud classification accuracy under various corruption scenarios, outperforming other state-of-the-art augmentation methods. Notably, under specific additive corruption types, our FFPS technique achieves an impressive error rate of 7.5\%, likely attributed to the weights utilized by FFPS that effectively filter out outliers. However, it is crucial to note that under the "\textit{Jitter}" corruption, PointSP does not attain optimal performance, where the PGD strategy excels, probably due to its unique optimization mechanism that better guides the model to learn robust features under subtle perturbations. Furthermore, for drop-type corruptions, the CutMix and Rsmix methods exhibit greater robustness, potentially because their strategies of mixing or replacing data segments enhance the model's tolerance to local or global information loss. Additional results can be found in the Appendix.

 \textbf{Results on PointCloud-C and OmniObject-C.} Table \ref{tab:merged_ER} compares the performance of PointSP with data augmentation methods on the PointCloud-C and OmniObject-C datasets. On PointCloud-C, PointSP enhances classification accuracy across various corruption scenarios, outperforming other methods, with the FFPS technique achieving an error rate of 7.5\% under additive corruption types due to its effective outlier filtering. However, PointSP does not perform optimally under the ``\textit{Jitter}" corruption, where the PGD strategy excels due to its robust feature learning mechanism. For drop-type corruptions, methods like CutMix and Rsmix demonstrate superior robustness, likely due to their data-mixing strategies. On OmniObject-C, PointSP excels in improving out-of-distribution (OOD) robustness, achieving the lowest mER for CurveNet (57.9\%) and the best results for ``\textit{Jitter}" (61.4\% mER). It also outperforms other methods on PointNet++ for ``\textit{Drop-G}" and ``\textit{Add-G}" corruptions, and is highly competitive with CutMix-R on PCT. Overall, these results validate PointSP’s effectiveness in enhancing both OOD robustness and generalization. Additional implementation details and results for PointNet and GDANet are available in the Appendix.

\begin{table}[t!] %\footnotesize %\scriptsize
\small
  \centering
  \setlength{\tabcolsep}{8.5pt}
  %\renewcommand\arraystretch{0.8}
  %\scalebox{0.85}{
  \begin{tabular}{l|c|ccc}
\toprule
Method  % (\%) %$\downarrow$     
      & mER         & Density       & Noise        & Transform         \\ 
\midrule \midrule
PointNet   & 28.3          & 28.3          & 32.7         & 24.0          \\
\ + CutMix-R  & 21.8          & 30.5          & \underline{18.0}         & 16.9          \\
\ + CutMix-K  & \underline{21.6}          & 26.8          & 21.8         & 16.3      \\
\ + Mixup      & 25.4          & 28.3          & 28.9         & 19.0          \\
\ + Rsmix      & 22.5          & \underline{24.8}    & 27.3         & \underline{15.5 }         \\
\ + PGD        & 25.9          & 28.8          & 28.4         & 20.5          \\ 
\ + \textbf{PointSP}       & \underline{21.6}          & 26.0          & 18.6         & 20.2          \\ 
\midrule
PointNet++ & 30.6          & 36.9          & 30.3        & 24.6       \\
\ + CutMix-R  & 19.8          & 26.8          & 14.0         & 18.6          \\
\ + CutMix-K  & 21.3          & 24.9          & 19.3         & 19.6          \\
\ + Mixup      & 18.6          & 29.7          & \underline{12.6}         & \underline{13.5}          \\
\ + Rsmix     & 27.0          & 28.9          & 23.8         & 28.3          \\ 
\ + \textbf{PointSP}       & \underline{17.7}          & \underline{\textbf{24.1}} & 12.7         & 16.2          \\ 
\midrule
GDANet     & 23.5          & 33.2          & 23.7         & 13.7          \\
\ + CutMix-R  & 16.9          & 28.5          & 10.4         & \underline{11.9}          \\
\ + CutMix-K  & 17.8          & 28.8          & 12.6         & \underline{11.9}          \\
\ + Mixup      & 18.5          & 30.3          & 13.1         & 12.2          \\
\ + Rsmix      & 19.2          & \underline{27.7}          & 14.4         & 15.4          \\
\ + PGD        & 20.3          & 32.1          & 15.9         & 13.0          \\ 
\ + \textbf{PointSP}       & \underline{16.8}          & 28.3          & \underline{10.0}   & 12.1          \\ 
\midrule
CurveNet   & 23.1          & 31.4          & 26.5         & \underline{11.4}          \\
\ + CutMix-R  & 16.1          & 25.7          & 10.5         & 12.1          \\
\ + CutMix-K  & 17.1          & \underline{24.8}    & 13.6         & 12.9          \\
\ + Mixup      & 20.8          & 32.4          & 17.9         & 12.1          \\
\ + Rsmix     & 19.9          & 26.7          & 15.6         & 17.3          \\
\ + PGD        & 20.4          & 28.5          & 21.3         & \underline{11.4}          \\ 
\ + \textbf{PointSP}       & \underline{\textbf{15.8}} & 25.6          & \underline{10.2}         & 11.7          \\ 
\midrule
PCT        & 25.5          & 34.8          & 28.1         & 13.5          \\
\ + CutMix-R  & 16.3    & 27.1          & 10.5         & 11.2    \\
\ + CutMix-K  & 16.5          & 25.8          & 12.6         & \underline{\textbf{11.1}} \\
\ + Mixup      & 19.5          & 30.3          & 16.7         & 11.5          \\
\ + Rsmix      & 17.3          & \underline{25.0}          & 12.0         & 15.0          \\
\ + PGD        & 18.4          & 29.3          & 14.7         & \underline{\textbf{11.1}} \\ 
\ + \textbf{PointSP}       & \underline{\textbf{15.8}} & 26.9          & \underline{\textbf{9.5}} & \underline{\textbf{11.1}} \\
\bottomrule
%    \toprule[1]
  \end{tabular} %}
  \caption{Comparison of mean error rate (mER) on ModelNet40-C between PointSP and state-of-the-art point cloud augmentation methods across five 3D deep models. The best mERs for each 3D deep model are underlined.}
  \label{tab:MNC_ER}
  \vspace{-1.5em}
\end{table}

\begin{table*}[htbp] %\scriptsize %\small
\centering
\setlength{\tabcolsep}{1.0pt}
\footnotesize
\begin{tabular}{l|c|ccccccc|c|ccccccc}
\toprule
\multirow{2}{*}{Method} & \multicolumn{8}{c|}{PointCloud-C} & \multicolumn{8}{c}{OmniObject-C} \\
\cmidrule(lr){2-9} \cmidrule(lr){10-17}
 & \textbf{mER} & Scale & Jitter & Drop-G & Drop-L & Add-G & Add-L & Rotate & \textbf{mER} & Scale & Jitter & Drop-G & Drop-L & Add-G & Add-L & Rotate \\
\midrule \midrule
PointNet++ & 27.7 & 9.4 & 50.3 & 26.2 & 39.6 & 15.9 & 20.2 & 32.5 & 73.9 & 64.9 & 80.7 & 77.9 & 78.5 & 71.6 & 71.1 & 72.4 \\
+ CutMix-R & 17.5 & 8.8 & 34.5 & \underline{\textbf{9.0}} & 20.9 & \underline{7.7} & \underline{8.1} & 33.7 & 64.9 & 62.9 & 72.2 & 62.8 & 67.2 & 59.3 & 58.1 & 71.8 \\
+ CutMix-K & 19.1 & 9.2 & 45.0 & 12.8 & 16.0 & 8.1 & 9.5 & 33.4 & 64.8 & 60.9 & 76.1 & 65.6 & 64.6 & 57.7 & 60.1 & 68.8 \\
+ Mixup & 17.7 & \underline{\textbf{8.5}} & \underline{25.2} & 16.4 & 27.2 & 9.5 & 11.7 & \underline{25.4} & 62.8 & \underline{59.3} & \underline{67.1} & 64.1 & 69.2 & 57.8 & \underline{57.2} & \underline{65.3} \\
+ Rsmix & 21.3 & 9.9 & 54.3 & 12.0 & \underline{14.3} & 7.9 & 8.9 & 41.9 & 66.7 & 63.8 & 78.1 & 66.2 & 68.1 & 59.4 & 60.6 & 70.6 \\
+ \textbf{PointSP} & \underline{17.3} & 9.2 & 33.3 & 10.9 & 16.0 & 8.0 & 10.3 & 33.1 & \underline{62.4} & 59.9 & 72.0 & \underline{58.6} & \underline{64.1} & \underline{56.8} & 57.8 & 67.7 \\
\midrule
CurveNet & 24.4 & 8.9 & 22.9 & 17.3 & 22.3 & 52.1 & 28.7 & 18.9 & 67.9 & 59.4 & 67.7 & 63.0 & 68.7 & 80.4 & 70.4 & 65.9 \\
+ CutMix-R & 13.8 & 9.1 & 18.2 & 11.1 & 15.5 & 8.1 & \underline{11.0} & 22.4 & 63.2 & 61.0 & 68.2 & 59.8 & 65.4 & 58.6 & 61.7 & 68.0 \\
+ CutMix-K & 15.8 & 8.7 & 30.6 & 12.9 & 10.3 & 8.5 & 15.5 & 23.9 & 60.3 & \underline{\textbf{57.3}} & 69.1 & 59.7 & \underline{\textbf{56.0}} & \underline{\textbf{53.2}} & 63.4 & \underline{\textbf{63.4}} \\
+ Mixup & 19.3 & \underline{8.6} & 17.9 & 21.6 & 19.8 & 25.7 & 20.1 & 21.4 & 62.5 & 57.7 & 62.7 & 60.9 & 65.7 & 64.1 & 61.7 & 64.6 \\
+ Rsmix & 16.9 & 9.1 & 35.0 & 11.0 & \underline{10.2} & 9.2 & 13.1 & 30.9 & 62.2 & 59.2 & 73.5 & 57.4 & 59.5 & 56.5 & 61.9 & 67.8 \\
+ PGD & 22.7 & 16.8 & \underline{11.2} & 12.9 & 26.0 & 48.9 & 25.3 & \underline{18.3} & 67.4 & 68.9 & 62.8 & 59.2 & 69.1 & 75.5 & 67.2 & 69.3 \\
+ \textbf{PointSP} & \underline{13.7} & 10.3 & 19.0 & \underline{10.3} & 11.0 & \underline{7.6} & 15.5 & 22.0 & \underline{57.9} & 58.1 & \underline{\textbf{61.4}} & \underline{\textbf{55.0}} & \underline{\textbf{56.0}} & 54.1 & \underline{\textbf{56.6}} & 64.0 \\
\midrule
PCT & 25.8 & \underline{9.0} & 27.1 & 15.0 & 24.1 & 40.3 & 42.9 & 22.2 & 69.8 & 59.3 & 71.3 & 60.4 & 68.7 & 83.0 & 80.6 & 65.5 \\
+ CutMix-R & 12.7 & 10.1 & 14.5 & 9.8 & 14.3 & 8.3 & 10.9 & 20.7 & \underline{60.8} & 59.5 & 62.9 & 58.8 & 60.8 & 57.7 & 61.2 & 64.7 \\
+ CutMix-K & 14.1 & 9.5 & 22.3 & 11.3 & 10.2 & 8.5 & 15.6 & 21.2 & 61.4 & \underline{\textbf{57.3}} & 65.8 & 62.8 & \underline{58.8} & \underline{56.3} & 65.1 & \underline{63.5} \\
+ Mixup & 18.1 & 9.4 & 15.6 & 15.8 & 18.2 & 23.5 & 22.8 & 21.1 & 62.7 & 57.6 & 62.0 & 58.9 & 63.6 & 65.7 & 67.5 & \underline{63.5} \\
+ Rsmix & 15.2 & 9.3 & 25.7 & 10.2 & \underline{\textbf{10.0}} & 8.7 & 13.0 & 29.8 & 63.3 & 59.2 & 70.5 & 60.6 & 59.8 & 58.6 & 65.9 & 68.5 \\
+ PGD & 20.0 & 14.6 & \underline{\textbf{10.5}} & 16.9 & 24.8 & 29.5 & 22.7 & 21.2 & 65.6 & 65.7 & \underline{61.5} & 66.7 & 71.8 & 62.6 & 64.3 & 66.7 \\
+WOLFMix & 12.7 & 9.4 & 27.0 & \underline{9.4} & 10.2 & 8.8 & 13.9 & \underline{\textbf{10.5}} & 60.5 & 59.3 & 61.9 & 59.1 & 60.7 & 58.2 & 60.1 & 64.5 \\
+ \textbf{PointSP} & \underline{\textbf{12.5}} & 9.9 & 16.2 & 11.0 & 14.4 & \underline{\textbf{7.5}} & \underline{\textbf{7.5}} & 21.1 & \underline{60.8} & 60.5 & 65.1 & \underline{58.1} & 60.9 & 57.7 & \underline{57.5} & 66.0 \\
\bottomrule
\end{tabular}
\caption{Comparison of mER on PointCloud-C and OmniObject-C datasets between PointSP and state-of-the-art point cloud enhancement methods across three 3D deep models.}
\label{tab:merged_ER}
\vspace{-1.5em}
\end{table*}

\subsection{Results for Part Segmentation}
\begin{figure}[h]
\centering
\includegraphics[width=0.48\textwidth]{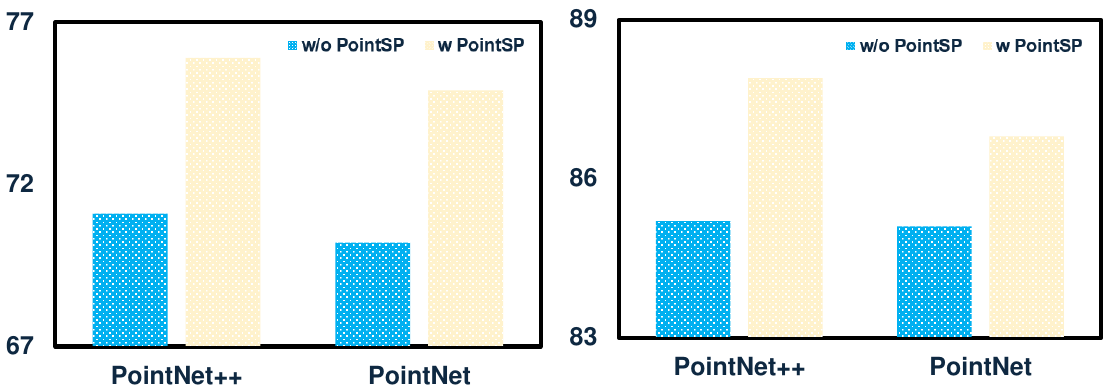}
\caption{ Instance mean IoU  (left) and overall accuracy (right) on ShapeNet-C.}
\label{fig:vis:shapeNet-C}
\end{figure}

The proposed sampling protocol has been evaluated on a classification task. To demonstrate its broader applicability, we also applied it to part segmentation tasks, which are critical for robotic manipulation, using the ShapeNet-C dataset \cite{ren2022pointcloud}. The results, shown in Figure \ref{fig:vis:shapeNet-C}, clearly indicate that the proposed PointSP protocol provides a significant improvement. 

% \begin{table}[] %\footnotesize
% \centering
% %\setlength{\tabcolsep}{1.0pt}
% %\tabcolsep=7pt
% %\renewcommand\arraystretch{0.9}
% %\scalebox{0.8}{
% \begin{tabular}{l|cc}
% \toprule
% Model & mIoU & OA  \\ 
% \midrule \midrule
% PointNet++ & 71.1 & 85.2 \\ 
% +PointSP & \textbf{75.9} & \textbf{87.9} \\\midrule
% PointNet & 70.2 & 85.1 \\
% +PointSP & \textbf{74.9} & \textbf{86.8} \\
% \bottomrule
% \end{tabular}
% \caption{Instance Average IoU (mIoU) and Overall Accuracy (OA) on ShapeNet-C}
% \label{tab:shapenet-c}
% \end{table}

\subsection{Ablation Studies}
In the ablation studies, we use PCT~\cite{guo2021pct} and CurveNet~\cite{xiang2021walk} as 3D deep models on two datasets: ModelNet40-C~\cite{sun2022benchmarking} and PointCloud-C (PCC)~\cite{ren2022pointcloud}. Further details on implementation and hyperparameters are provided in Appendix.

\begin{table}[] 
\footnotesize
\centering
\setlength{\tabcolsep}{4.5pt}
\renewcommand{\arraystretch}{0.9}
\begin{tabular}{c|ccc|cc|cc}
\toprule
\multicolumn{1}{c|}{\begin{tabular}[c]{@{}c@{}}Full Points\\ Resampling\end{tabular}} & \multicolumn{3}{c|}{Key Points Sampling} & \multicolumn{2}{c|}{PCT}    & \multicolumn{2}{c}{CurveNet}  \\ 
\midrule
   \textbf{RUD}    & FPS    & FFPS   &   \textbf{SWS}   & MNC           & PCC           & MNC           & PCC           \\ 
\midrule \midrule
    &        &   &  \checkmark  & 19.2          & 14.8          & 22.7          & 20.1          \\
   \checkmark    &\checkmark  &           &       & 16.8        & 13.3        & 16.3      & 14.1    \\
\checkmark   &         &\checkmark   &          & 17.0        & 13.3        & 16.0        & 13.8   \\
\checkmark   &      &    & \checkmark    & \textbf{15.8} & \textbf{12.5} & \textbf{15.8} &\textbf{13.7} \\
\bottomrule
\end{tabular}

%}
\caption{Ablation study based on mER with different sampling protocols \textbf{in training}. RUD: %refers to 
random upsampling \& downsampling; FFPS: %represents the 
filtered FPS; SWS: %$\mathbf{s}_t \sim$  Cat($\cdot$) 
% is the 
stochastic weighted sampling. %\color{red}{cline problem.} 
}%. as Equ (\ref{eq:training:downsampling}). } %ablation study in the training phase }
\label{tab:train_ablation}
% \vspace{-1.5em}
\end{table}

\textbf{Sampling Protocols in Training.} We compared various sampling protocols during the training phase. As shown in Table \ref{tab:train_ablation}, the combination of random up\&downsampling (RUD) and stochastic weighted sampling (SWS) consistently delivers the best performance. Removing RUD or substituting the proposed SWS method with FPS obviously degrades the performance. FFPS also achieves the second-based performance, highlighting the importance of point reweighting.

\textbf{Random Upsampling and Downsampling.} We investigate the impact of various resampling techniques during training. As shown in Table \ref{tab:ablation_upsampling}, we explore different methods for increasing and reducing sample sizes. The results indicate that the proposed local-glocal-balanced method plays a crucial role in enhancing performance. This suggests that stochastically determining the localness of the point dropping, \textit{i.e,} neighbor size, can improve robustness against corruptions. Additionally, we compare our upsampling method in Algorithm \ref{alg:estimate_perb} with Shape-invariant perturbation (SI), which conducts per-point perturbation on the tangent plane \cite{huang2022shape}. The superiority of our method over SI shows that perturbation direction based on neighbor points preserves more local information than random directions. 

\begin{table}[]
\small
\centering
\setlength{\tabcolsep}{1.8pt}
\begin{tabular}{cc|ccc|cc|cc}
\toprule
\multicolumn{2}{c|}{$\Delta N >0$} & \multicolumn{3}{c|}{$\Delta N  \leqslant 0$} & \multicolumn{2}{c|}{PCT}    &\multicolumn{2}{c}{CurveNet}  \\
\midrule
SI          & \textbf{LGP}          & RD     & KNN     & \textbf{LGB}     & MNC           & PCC           & MNC           & PCC           \\ 
\midrule \midrule
\checkmark  &             &            &         &\checkmark & \textbf{15.8} & 12.6 & 17.2          & 14.7          \\
            & \checkmark  & \checkmark &         &           & 17.0          & 13.4          & 16.7          & 14.8          \\
            &\checkmark   &            & \checkmark &        & 17.2          & 13.8          & 17.4          & 15.9          \\
            & \checkmark  &            &         &\checkmark & \textbf{15.8} & \textbf{12.5}          & \textbf{15.8} & \textbf{13.7} \\ 
\bottomrule
\end{tabular}
\caption{Ablation study based on mER of \textbf{full points resampling} using different techniques for upsampling ($\Delta N >0$) and removing points ($\Delta N \leqslant 0$). SI: %represents 
shape-invariant tangent plane upsampling with random directions and distances; %\cite{huang2022shape}; 
\textbf{LGP}: %denotes the proposed 
the proposed local-geometry-preserved upsampling; % (proposed); %detailed in Algorithm \ref{alg:estimate_perb}. 
RD: %refers to the 
random global downsampling of points;  KNN: local neighbours removal;
\textbf{LGB}: %represents 
the proposed local-gobal balanced downsampling. %strategy. 
}
\label{tab:ablation_upsampling}
\vspace{-1.0em}
\end{table}

\begin{figure}[]
\centering
\subfigure{
\includegraphics[width=4.2cm]{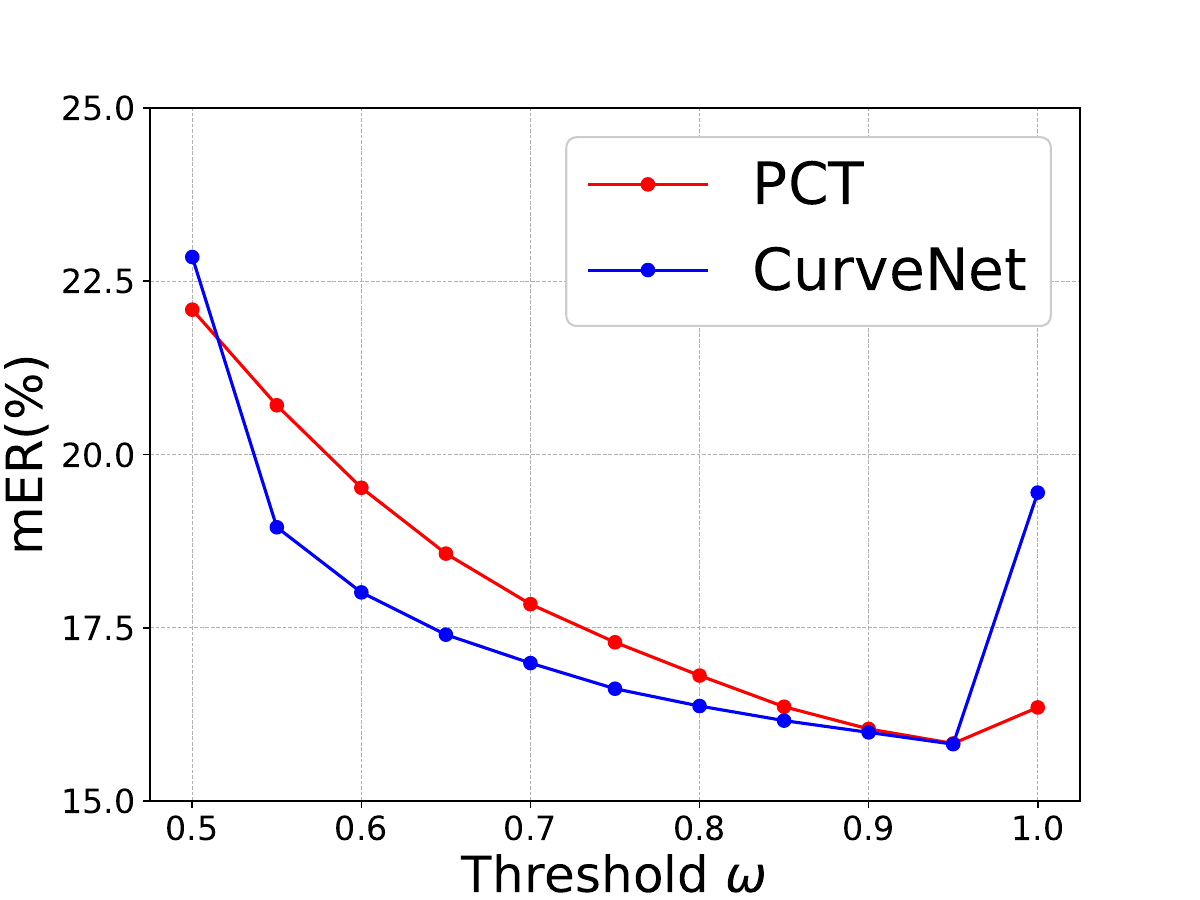}
\label{fig:mnc-ffps}}\subfigure{
\includegraphics[width=4.2cm]{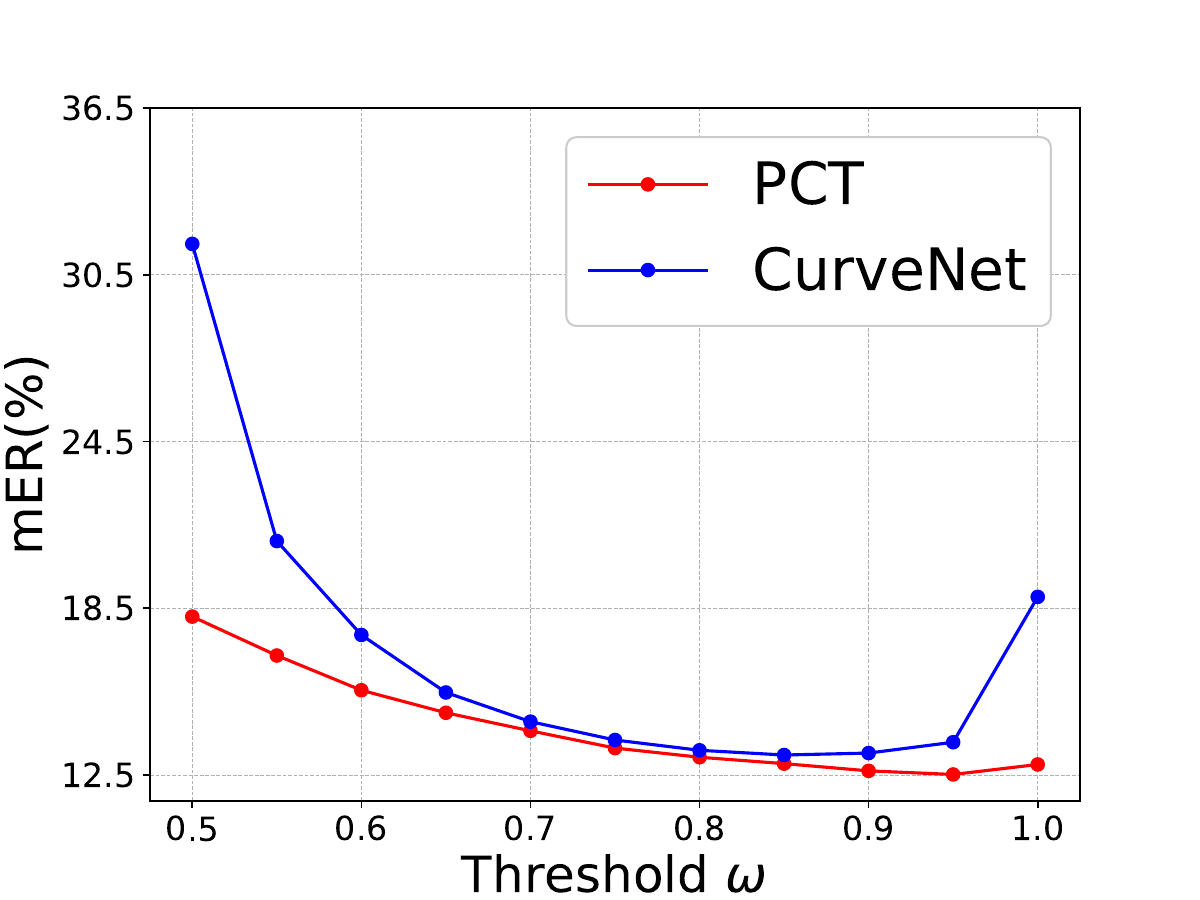}
\label{fig:pcc-ffps}
}
\vspace{-1.2em}
\caption{mERs of FFPS's different threshold $\omega$ on ModelNet40-C (left) and PointCloud-C (right).}
\label{fig:w-fps-omega}
\vspace{-1em}
\end{figure}

\textbf{Effect of Quantile-based Threshold  $\omega$ of FFPS.} As illustrated in Figure \ref{fig:w-fps-omega}, retain 95\% of the points (i.e., filtering out the 5\% of points with the smallest weights) results in the best performance. Moreover, when nearly half of the points are removed, the mER peaks, likely due to the loss of critical information within the point cloud. As the number of removed points decreases, the error rate also decreases, reaching its lowest when about 5\% of the points are filtered out, before rising again. It suggests that an optimal balance is achieved.

\subsection{Visualization Results}

\begin{figure}[]
\centering
\includegraphics[width=0.47\textwidth]{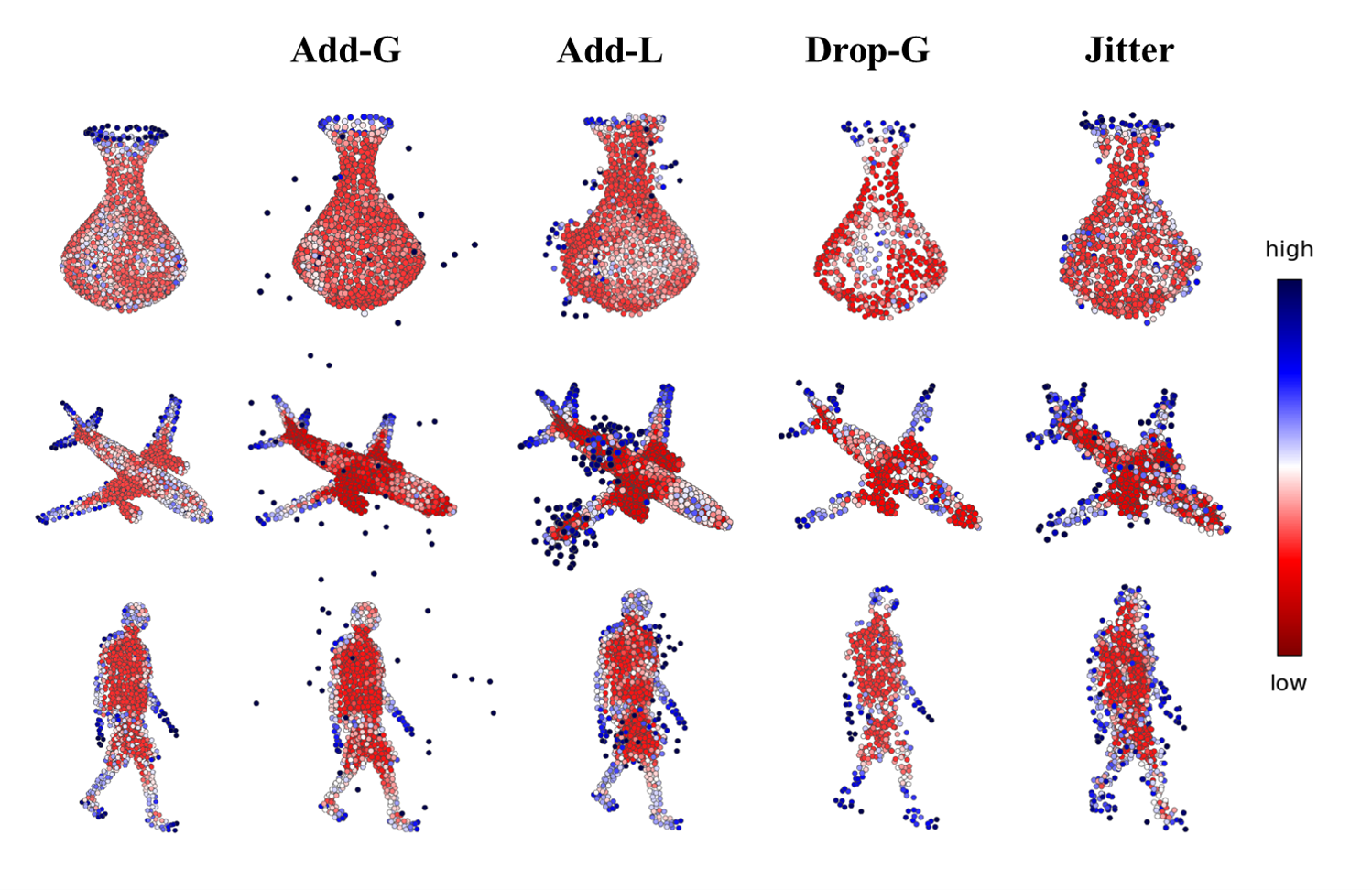}
\caption{Visualization of point-wise \textit{Isolation Rate}. Lower isolation rate Column 1 presents the clean data, while Columns 2 to 5 depict data with the corresponding corruption types indicated above each column.}
\label{fig:vis:point-weights}
\end{figure}

\textbf{Isolation Rate.} In Figure \ref{fig:vis:point-weights}, we visualize the distribution of point-wise isolation rates for three example objects. The proposed rate effectively identifies boundary points and outliers, thereby enhancing subsequent point cloud sampling and improving learning robustness against corruption. 

\begin{figure}[]
\centering
\includegraphics[width=0.44\textwidth]{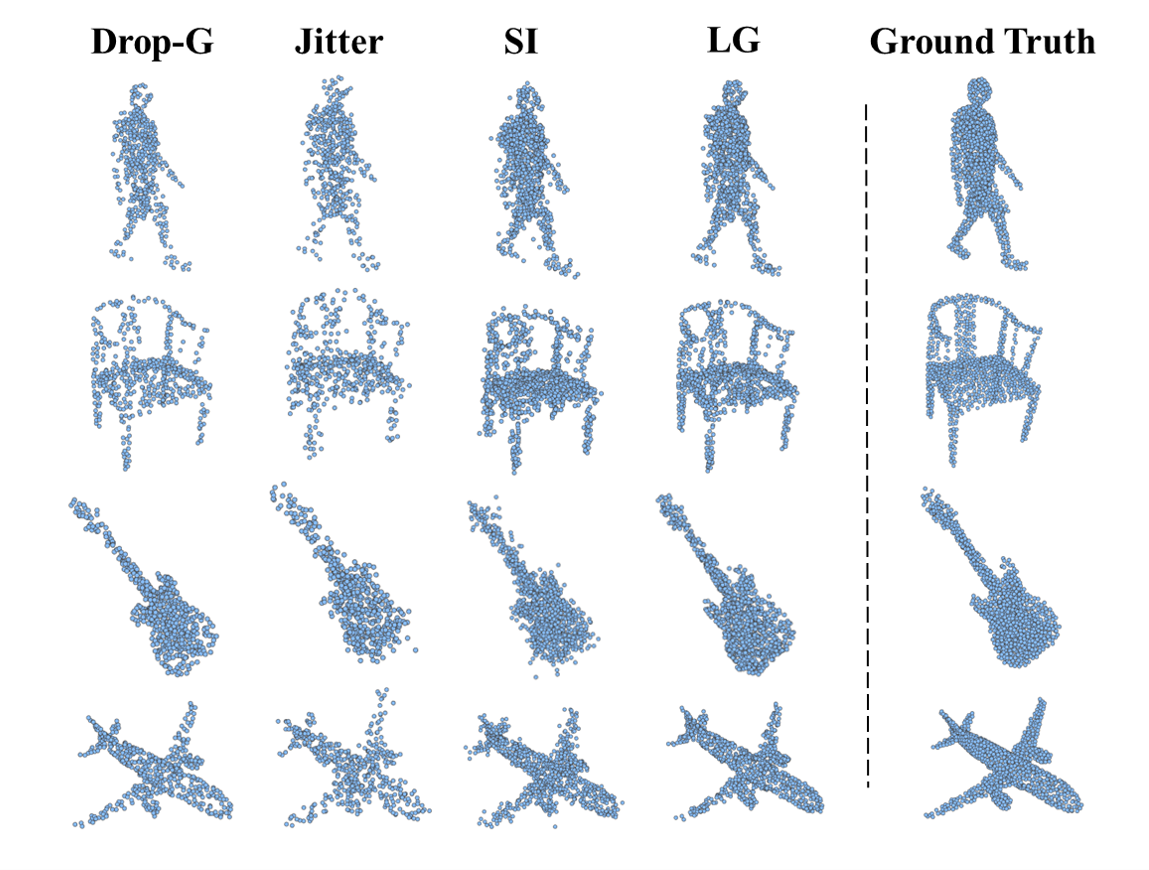}
\caption{
Visual comparison of three sampling techniques: (1) Jitter, (2) SI: tangent plane upsampling with random directions and distances %~\cite{huang2022shape}, 
and (3) LG: the proposed local-geometry-preserved upsampling. The first column shows the corrupted data from PointCloud-C, while the last column presents the corresponding clean data. }
\label{fig:vis:upsampling}
\end{figure}

\textbf{Local-geometry-preserved Interpolation.} Figure \ref{fig:vis:upsampling} visually compares the results of three upsampling techniques on four example objects. It is evident that both Jitter and SI~\cite{huang2022shape} struggle with corrupted data, particularly when it is sparse and non-uniform. In contrast, the proposed LGP method effectively combines completion and uniformity in the upsampling process. 

\begin{figure}[]
\centering
\includegraphics[width=0.47\textwidth]{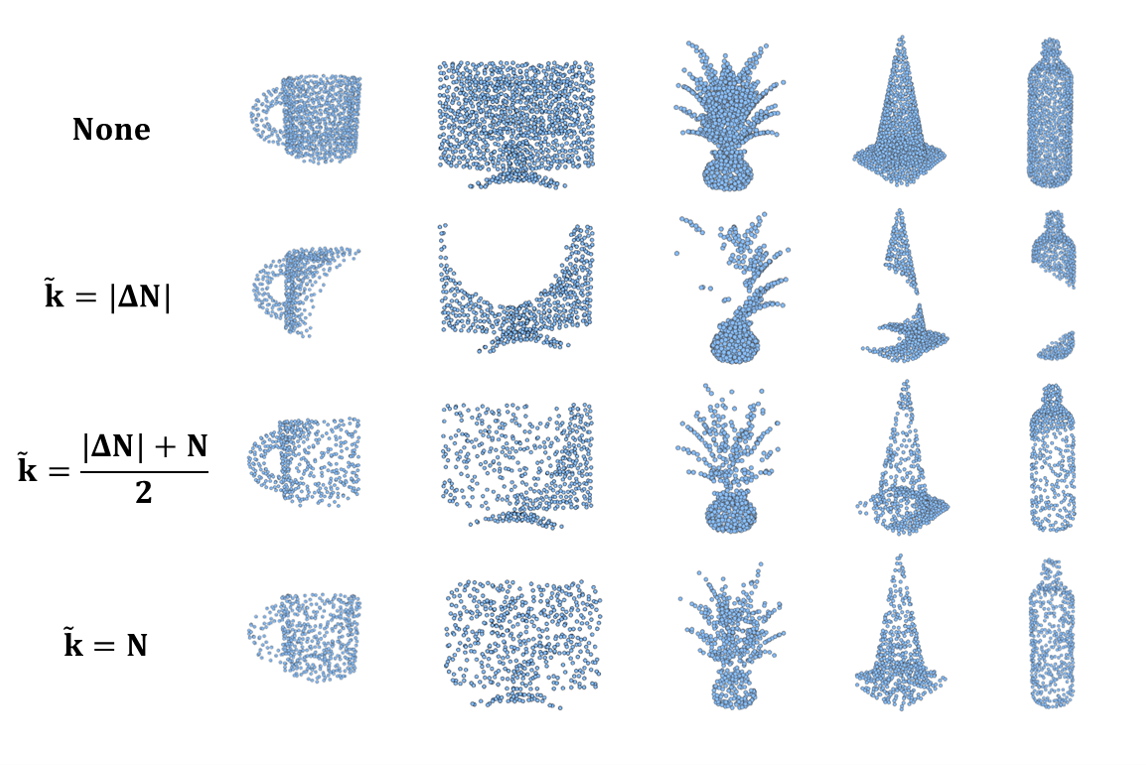}
\caption{Visualization of samples with different neighborhood size $\Tilde{k}$ in local-global-balanced downsampling.}
\label{fig:vis:global_score}
\end{figure}

\textbf{Neighborhood Size $\Tilde{k}$ in Local-global-balanced downsampling.} Stochastically determining the sample size is a critical aspect of the resampling protocol. %, driven by the stochastic neighborhood size $\Tilde{k}$. 
As shown in Figure \ref{fig:vis:global_score}, a smaller $\Tilde{k}$ leads to local drops (second row), while a larger $\Tilde{k}$ results in more global removals (last row). A stochastic $\Tilde{k}$ would closely mimic real-world corruption, contributing to the robust improvement of the proposed protocol.

%\vspace{-1mm}
\section{Conclusion} \label{sec:con}
This work highlights the limitations of current sampling protocols for corrupted 3D point clouds and proposes PointSP, a robust solution that mitigates outliers and restores incomplete clouds without additional training. Experiments show that PointSP outperforms state-of-the-art methods in 3D classification, paving the way for more reliable 3D deep learning in critical applications like autonomous driving. 
%This work focuses on the safety issue of 3D point cloud deep learning. It indicates that the current sampling protocol is not optimized for corrupted 3D point cloud analysis, thus posing a potential threat to 3D applications. Therefore, we introduce PointSP, an enhanced sampling protocol for point clouds, addressing shortcomings in handling real-world corruptions. By incorporating point reweighting and tangent plane interpolation, our protocol mitigates outliers and restores incomplete point clouds without additional training or architectural changes. %Extensive experiments show PointSP significantly improves robustness in 3D classification, outperforming state-of-the-art methods. This approach paves the way for more reliable 3D deep learning in critical areas like autonomous driving. Future research could extend this protocol to 3D scene-based datasets and higher-level tasks.

\bibliographystyle{named}
\bibliography{RobustSampling}

\end{document}